\documentclass[10pt,onecolumn,letterpaper, twocolumn]{article}

\usepackage{cvpr}
\usepackage{times}

\usepackage{color,xcolor}
\usepackage{epsfig}
\usepackage{graphicx}

\usepackage{adjustbox}
\usepackage{array}
\usepackage{booktabs}
\usepackage{colortbl}
\usepackage{float,wrapfig}
\usepackage{hhline}
\usepackage{multirow}
\usepackage{subcaption} 
\usepackage[toc,page]{appendix}

\usepackage{amsmath,amsfonts,amsthm,amssymb}
\usepackage{bm}
\usepackage{nicefrac}
\usepackage{microtype}

\usepackage{changepage}
\usepackage{extramarks}
\usepackage{fancyhdr}
\usepackage{lastpage}
\usepackage{setspace}
\usepackage{soul}
\usepackage{xspace}
\usepackage{indentfirst}

\usepackage[pagebackref=true,breaklinks=true,letterpaper=true,colorlinks,bookmarks=false]{hyperref}
\usepackage{url}

\usepackage{algorithm, algorithmic}
\usepackage{enumerate}
\newcolumntype{L}[1]{>{\raggedright\let\newline\\\arraybackslash\hspace{0pt}}m{#1}}
\newcolumntype{C}[1]{>{\centering\let\newline\\\arraybackslash\hspace{0pt}}m{#1}}
\newcolumntype{R}[1]{>{\raggedleft\let\newline\\\arraybackslash\hspace{0pt}}m{#1}}

\newcommand{\sect}[1]{Section~\ref{#1}}

\newcommand{\fig}[1]{Figure~\ref{#1}}

\newcommand{\tbl}[1]{Table~\ref{#1}}

\newcommand{\ignorethis}[1]{}

\DeclareMathOperator*{\argmin}{arg\,min}

\makeatletter
\DeclareRobustCommand\onedot{\futurelet\@let@token\@onedot}
\def\@onedot{\ifx\@let@token.\else.\null\fi\xspace}

\def\eg{\emph{e.g}\onedot} 
\def\ie{\emph{i.e}\onedot} 
 
\def\etc{\emph{etc}\onedot} 
 
\def\etal{\emph{et al}\onedot}
\makeatother

\newcommand{\myparagraph}[1]{\vspace{-3pt}\paragraph{#1}}

\def\model{\textbf{H}ardware-Aware \textbf{A}utomated \textbf{Q}uantization\xspace}
\def\modelshort{HAQ\xspace}

\cvprfinalcopy

\begin{document}

\title{HAQ: Hardware-Aware Automated Quantization with Mixed Precision}
\author{
    Kuan Wang$^*$, Zhijian Liu$^*$, Yujun Lin$^*$, Ji Lin, and Song Han \\
    {\tt\small\{kuanwang, zhijian, yujunlin, jilin, songhan\}@mit.edu} \\
    Massachusetts Institute of Technology
}

\maketitle

\footnotetext{$*$ indicates equal contributions.}

\begin{abstract}

Model quantization is a widely used technique to compress and accelerate deep neural network (DNN) inference. Emergent DNN hardware accelerators begin to support \emph{mixed precision} (1-8 bits) to further improve the computation efficiency, which raises a great challenge to find the optimal bitwidth for each layer: it requires domain experts to explore the vast design space trading off among accuracy, latency, energy, and model size, which is both time-consuming and sub-optimal. There are plenty of specialized hardware for neural networks, but little research has been done for specialized neural network optimization for a particular hardware architecture. Conventional quantization algorithm ignores the different hardware architectures and quantizes all the layers in a uniform way. In this paper, we introduce the \model (\modelshort) framework which leverages the reinforcement learning to automatically determine the quantization policy, and we take the hardware accelerator's feedback in the design loop. Rather than relying on proxy signals such as FLOPs and model size, we employ a hardware simulator to generate direct feedback signals (latency and energy) to the RL agent. Compared with conventional methods, our framework is fully automated and can specialize the quantization policy for different neural network architectures and hardware architectures. Our framework effectively reduced the latency by \textbf{1.4-1.95$\times$} and the energy consumption by \textbf{1.9$\times$} with negligible loss of accuracy compared with the fixed bitwidth (8 bits) quantization. Our framework reveals that the optimal policies on different hardware architectures (\ie, edge and cloud architectures) under different resource constraints (\ie, latency, energy and model size) are drastically different. We interpreted the implication of different quantization policies, which offer insights for both neural network architecture design and hardware architecture design.

\end{abstract}

\section{Introduction}
\label{sec:intro}

In many real-time machine learning applications (such as robotics, autonomous driving, and mobile VR/AR), deep neural networks is strictly constrained by the latency, energy, and model size. In order to improve the hardware efficiency, many researchers have proposed to quantize the weights and activations to low precision~\cite{Han:2016uf,Lin:2017ww,Zhu:2017wy}.

\begin{figure}[!t]
    \centering
    \includegraphics[width=\linewidth]{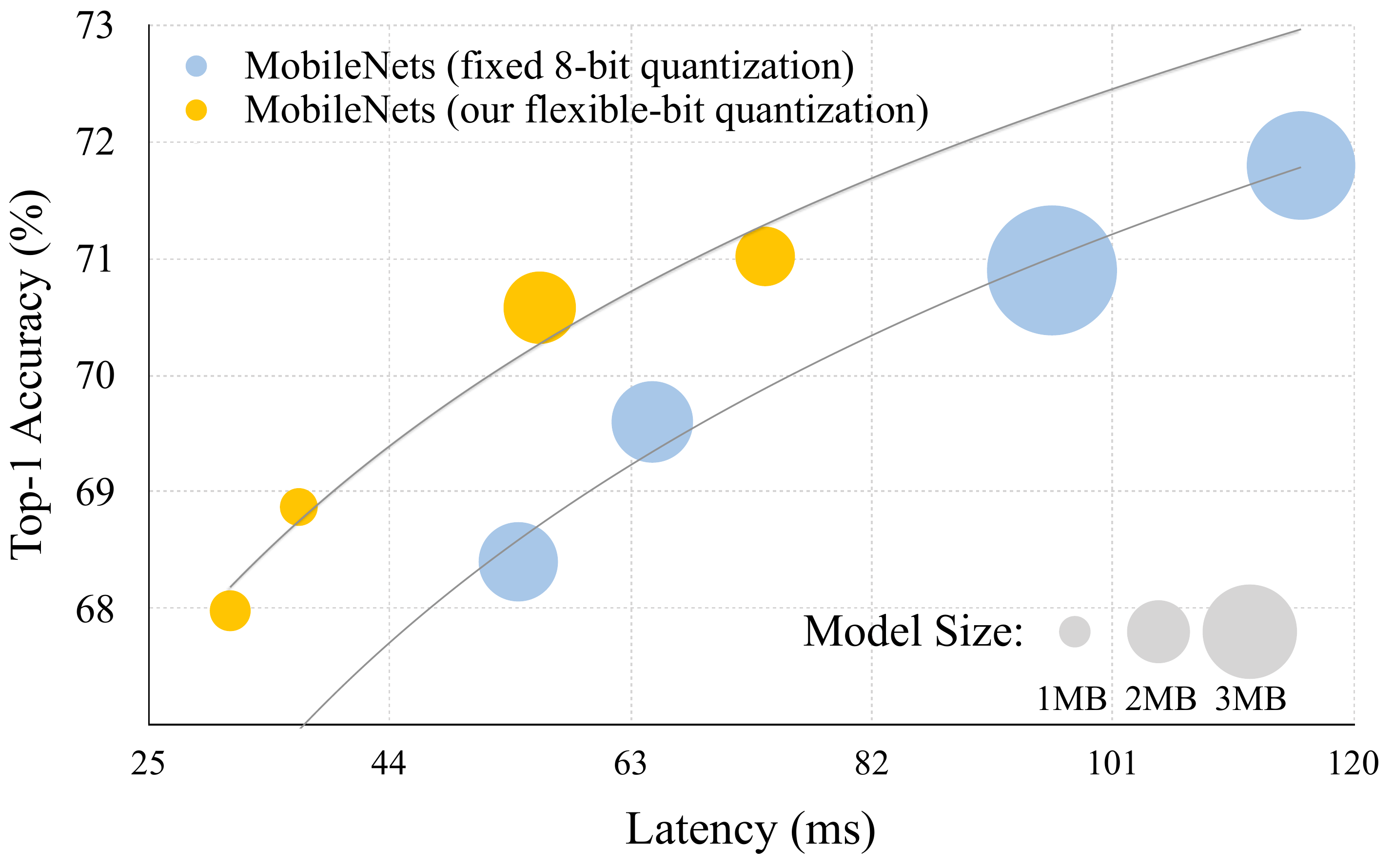}
    \caption{We need \textbf{mixed precision} for different layers. We quantize MobileNets~\cite{Howard:2017wz} to different number of bits (both weights and activations), and it lies on a better pareto curve (yellow) than fixed bit quantization (blue). The reason is that different layers have different redundancy and have different arithmetic intensity (OPs/byte) on the hardware, which advocates for using mixed precision for different layers.}
    \label{fig:teaser}
    \vspace{-12pt}
\end{figure}

Conventional quantization methods use the same number of bits for all layers~\cite{Choi:2018uw,Jacob:2018ur}, but as different layers have different redundancy and behave differently on the hardware (computation bounded or memory bounded), it is necessary to use \emph{mixed precision} for different layers (as shown in \fig{fig:teaser}). This flexibility was originally not supported by chip vendors until recently the hardware manufacturers started to implement this feature: Apple released the A12 Bionic chip that supports mixed precision for the neural network inference~\cite{apple}; NVIDIA recently introduced the Turing GPU architecture that supports 1-bit, 4-bit, 8-bit and 16-bit arithmetic operations~\cite{nvidia}; Imagination launched a flexible neural network IP that supports per-layer bitwidth adjustment for both weights and activations~\cite{imagination}. Besides industry, recently academia also works on the bit-level flexible hardware design: BISMO~\cite{umuroglu2018bismo} proposed the bit-serial multiplier to support multiplications of 1 to 8 bits; BitFusion~\cite{sharma2018bit} supports multiplications of 2, 4, 8 and 16 bits in a spatial manner.

\begin{table}[!t]
    \renewcommand*{\arraystretch}{1.4}
    \small\centering
    \begin{tabular}{lc|c|c}
        \toprule
        & \multicolumn{3}{c}{Inference latency on} \\
        & \textbf{HW1} & \textbf{HW2} & \textbf{HW3} \\
        \midrule
        Best Q. policy for \textbf{HW1} & \cellcolor{red!15}\textbf{16.29} ms & 85.24 ms & 117.44 ms \\
        Best Q. policy for \textbf{HW2} & 19.95 ms & \cellcolor{red!15}\textbf{64.29} ms & 108.64 ms \\
        Best Q. policy for \textbf{HW3} & 19.94 ms & 66.15 ms & \cellcolor{red!15}\textbf{99.68} ms \\
        \bottomrule
    \end{tabular}
    \vspace{-6pt}
    \caption{Inference latency of MobileNet-V1~\cite{Howard:2017wz} on three hardware architectures under different quantization policies. The quantization policy that is optimized for one hardware is not optimal for the other. This suggests we need a \textbf{specialized} quantization solution for different hardware architectures. (HW1: BitFusion \cite{sharma2018bit}, HW2: BISMO \cite{umuroglu2018bismo} edge accelerator, HW3: BISMO cloud accelerator, batch = 16).}
    \label{tbl:teaser}
    \vspace{-12pt}
\end{table}

However, a very missing part is how to \textbf{determine the bitwidth of both weights and activations for each layer on different hardware accelerators}. This is a vast design space: with $M$ different neural network models, each with $N$ layers, on $H$ different hardware platforms, there are in total $\mathcal{O}(H \times M \times 8^{2N})$\footnote{Assuming the bitwidth is 1 to 8 for both weights and activations.} possible solutions. For a widely used ResNet-50~\cite{He:2015tt} model, the size of the search space is about $8^{100}$, which is even larger than the number of particles in the universe. Conventional methods require domain experts (with knowledge of both machine learning and hardware architecture) to explore the huge design space smartly with rule-based heuristics, such as: we should retain more bits in the first layer which extracts low level features and in the last layer which computes the final outputs; also, we should use more bits in the convolution layers than in the fully-connected layers because empirically, the convolution layers are more sensitive. As the neural network becomes deeper, the search space increases exponentially, which makes it infeasible to rely on hand-crafted strategies. Therefore, these \emph{rule-based} quantization policies are usually sub-optimal, and they cannot generalize from one model to another. In this paper, we would like to \emph{automate} this exploration process by a \emph{learning-based} framework.

Another  challenge is how to optimize the latency and the energy consumption of a given model on the hardware. A widely adopted approach is to rely on some proxy signals (\eg, FLOPs, number of memory references)~\cite{Howard:2017wz,Sandler:2018wy}. However, as different hardware behaves very differently, the performance of a model on the hardware cannot always be accurately reflected by these proxy signals. Therefore, it is important to directly \emph{involve the hardware architecture's performance feedback into the design loop}. Also, as demonstrated in \tbl{tbl:teaser}, the quantization solution optimized on one hardware  might not be optimal on the other, which raises the demand for \emph{specialized} policies for different hardware architectures.

To this end, we propose the \model (\modelshort) framework that leverages reinforcement learning to automatically predict the quantization policy given the hardware's feedback. 
The RL agent decides the bitwidth of a given neural network in a layer-wise manner. For each layer, the agent receives the layer configuration and statistics as observation, and it then outputs the action which is the bitwidth of weights and activations. We then leverage the hardware accelerator as the environment to obtain the \emph{direct feedback from hardware} to guide the RL agent to satisfy the resource constraints. After all layers are quantized, we finetune the quantized model for one more epoch, and feed the validation accuracy after short-term retraining as the reward signal to our RL agent. During the exploration, we leverage the deep deterministic policy gradient (DDPG)~\cite{Lillicrap:2016ww} to supervise our RL agent. We also studied the quantization policy on multiple hardware architectures: both cloud and edge neural network accelerators, with spatial or temporal multi-precision design. 

The contribution of this paper has four aspects:
\begin{enumerate}
    \item \textbf{Automation}: We propose an automated framework for quantization, which does not require domain experts and rule-based heuristics. It frees the human labor from exploring the vast search space of choosing bitwidths.
    \item \textbf{Hardware-Aware}: Our framework involves the hardware architecture into the loop so that it can directly reduce the latency, energy and storage on the target hardware instead of relying on proxy signals.
    \item \textbf{Specialization}: For different hardware architectures, our framework can offer a specialized quantization policy that's exactly tailored for the target hardware architecture to optimize latency and energy.
    \item \textbf{Design Insights}: We interpreted the different quantization polices learned for different hardware architectures. Taking both computation and memory access into account, the interpretation offers insights on both neural network architecture and hardware architecture design.
\end{enumerate}

\section{Related Work}

\begin{figure*}[t]
    \centering
    \includegraphics[width=0.93\textwidth]{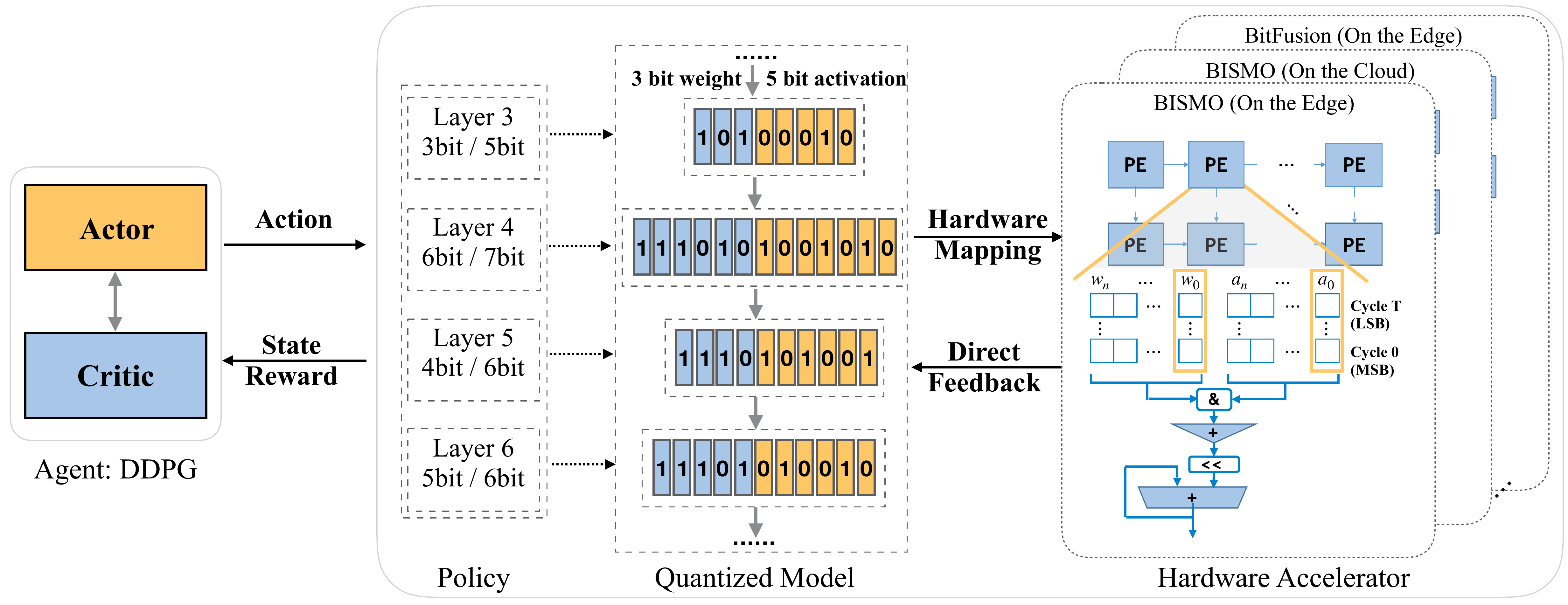}
    \caption{An overview of our \model (\modelshort) framework. We leverage the reinforcement learning to automatically search over the huge quantization design space with hardware in the loop. The agent propose an optimal bitwidth allocation policy given the amount of computation resources (\ie, latency, power, and model size). Our RL agent integrates the hardware accelerator into the exploration loop so that it can obtain the direct feedback from the hardware, instead of relying on indirect proxy signals.}
    \label{fig:overview}
    \vspace{-12pt}
\end{figure*}

\paragraph{Quantization.}

There have been extensive explorations on compressing and accelerating deep neural networks using quantization. Han~\etal~\cite{Han:2016uf} quantized the network weights to reduce the model size by rule-based strategies: \eg, they used human heuristics to determine the bitwidths for convolution and fully-connected layers. Courbariaux~\etal~\cite{Courbariaux:2016tm} binarized the network weights into $\{-1, +1\}$; Rastegari~\etal~\cite{Rastegari:2016tn} and Zhou~\etal~\cite{zhou2018explicit} binarized each convolution filter into $\{-w, +w\}$; Zhu~\etal~\cite{Zhu:2017wy} mapped the network weights into $\{-w_\text{N}, 0, +w_\text{P}\}$ using two bits; Zhou~\etal~\cite{Zhou:2016wh} used one bit for network weights and two bits for activations; Jacob~\etal~\cite{Jacob:2018ur} made use of 8-bit integers for both weights and activations. We refer the reader to the survey paper by Krishnamoorthi~\etal~\cite{Krishnamoorthi:2018wr} for a more detailed overview. These conventional quantization methods either simply assign the same number of bits to all layers or require domain experts to determine the bitwidths for different layers, while our framework automates this design process, and our \emph{learning-based} policy outperforms \emph{rule-based} strategies.

\myparagraph{AutoML.}

Many researchers aimed to improve the performance of deep neural networks by searching the network architectures: Zoph~\etal~\cite{Zoph:2017uo} proposed the Neural Architecture Search (NAS) to explore and design the transformable network building blocks, and their network architecture outperforms several human designed networks; Liu~\etal~\cite{Liu:2018tr} introduced the Progressive NAS to accelerate the architecture search by 5$\times$ using sequential model-based optimization; Pham~\etal~\cite{Pham:2018tl} introduced the Efficient NAS to speed up the exploration by 1000$\times$ using parameter sharing; Cai~\etal~\cite{Cai:2018wb} introduced the path-level network transformation to effectively search the tree-structured architecture space. Motivated by these AutoML frameworks, He~\etal~\cite{He:2018vj} leveraged the reinforcement learning to automatically prune the convolution channels. Our framework further explores the automated quantization for network weights and activations, and it takes the hardware architectures into consideration.

\myparagraph{Efficient Models.}

To facilitate the efficient deployment, researchers designed hardware-friendly approaches to slim neural network models. For instance, the coarse-grained channel pruning methods~\cite{he2017channel, liu2017learning} prune away the entire channel of convolution kernels to achieve speedup. 
Recently, researchers have explicitly optimized for various aspects of hardware properties, including the inference latency and energy: Yang~\etal~\cite{yang2016designing} proposed the energy-aware pruning to directly optimize the energy consumption of neural networks; Yang~\etal~\cite{yang2018netadapt} reduced the inference time of neural networks on the mobile devices through a lookup table.
Nevertheless, these methods are still rule-based and mostly focus on pruning. Our framework automates the quantization process by taking hardware-specific metric as direct rewards using a learning based method.
\section{Approach}

We model the quantization task as a reinforcement learning problem (\fig{fig:overview}). We use the actor-critic model with DDPG agent to give the action: bits for each layer. We collect hardware counters as constraints, together with accuracy as rewards to search the optimal quantization policy. We have three hardware environments that covers edge and cloud, spatial and temporal architectures for mixed-precision accelerator. Below describes the details of the RL formulation.

\subsection{Observation (State Space)}

Our agent processes the neural network in a layer-wise manner. For each layer, our agent takes two steps: one for weights, and one for activations. In this paper, we introduce a ten-dimensional feature vector ${O}_k$ as our observation:

\vspace{4pt}\noindent
If the $k$\textsuperscript{th} layer is a convolution layer, the state ${O}_k$ is 
\begin{equation}
    {O}_k = (k, c_\text{in}, c_\text{out}, s_\text{kernel}, s_\text{stride}, s_\text{feat}, n_\text{params}, i_\text{dw}, i_\text{w/a}, a_{k - 1}),
\end{equation}
where $k$ is the layer index, $c_\text{in}$ is \#input channels, $c_\text{out}$ is \#output channels, $s_\text{kernel}$ is kernel size, $s_\text{stride}$ is the stride, $s_\text{feat}$ is the input feature map size, $n_\text{params}$ is \#parameters, $i_\text{dw}$ is a binary indicator for depthwise convolution, $i_\text{w/a}$ is a binary indicator for weight/activation, and $a_{k - 1}$ is the action from the last time step.

\vspace{4pt}\noindent
If the $k$\textsuperscript{th} layer is a fully-connected layer, the state ${O}_k$ is
\begin{equation}
    {O}_k = (k, h_\text{in}, h_\text{out}, 1, 0, s_\text{feat}, n_\text{params}, 0, i_\text{w/a}, a_{k-1}),
\end{equation}
where $k$ is the layer index, $h_\text{in}$ is \#input hidden units, $h_\text{out}$ is \#output hidden units, $s_\text{feat}$ is the size of input feature vector, $n_\text{params}$ is \#parameters, $i_\text{w/a}$ is a binary indicator for weight/ activation, and $a_{k - 1}$ is the action from the last step.

For each dimension in the observation vector ${O}_k$, we normalize it into $[0, 1]$ to make them in the same scale.

\subsection{Action Space}
\label{sect:approach:action_space}

We use a \emph{continuous} action space to determine the bitwidth. The reason that we do not use a \emph{discrete} action space is because it loses the relative order: \eg, 2-bit quantization is more aggressive than 4-bit and even more than 8-bit. At the $k$\textsuperscript{th} time step, we take the continuous action $a_k$ (which is in the range of $[0, 1]$), and round it into the discrete bitwidth value $b_k$:
\begin{equation}
    b_k = \text{round} (b_\text{min} - 0.5 + a_k \times (b_\text{max} - b_\text{min} + 1)),
\end{equation}
where $b_\text{min}$ and $b_\text{max}$ denote the min and max bitwidth (in our experiments, we set $b_\text{min}$ to $2$ and $b_\text{max}$ to $8$).

\myparagraph{Resource Constraints.}

In real-world applications, we have limited computation budgets (\ie, latency, energy, and model size). We would like to find the quantization policy with the best performance given the constraint.

We encourage our agent to meet the computation budget by limiting the action space. After our RL agent gives actions $\{a_k\}$ to all layers, we measure the amount of resources that will be used by the quantized model. The feedback is directly obtained from the hardware accelerator, which we will discuss in \sect{sect:hardware_accelerator}. If the current policy exceeds our resource budget (on latency, energy or model size), we will sequentially decrease the bitwidth of each layer until the constraint is finally satisfied.

\subsection{Direct Feedback from Hardware Accelerators}
\label{sect:hardware_accelerator}

An intuitive feedback to our RL agent can be FLOPs or the model size. However, as these proxy signals are indirect, they are not equal to the performance (\ie, latency, energy consumption) on the hardware.  Cache locality, number of kernel calls, memory bandwidth all matters. Proxy feedback can not model these hardware functionality to find the specialized strategies (see \tbl{tbl:teaser}).

Instead, we use direct latency and energy feedback from the hardware accelerator as resource constraints, which enables our RL agent to determine the bitwidth allocation policy from the subtle differences between different layers: \eg, vanilla convolution has more data reuse and better locality, while depthwise convolution~\cite{Chollet:2017vb} has less reuse and worse locality, which makes it memory bounded. Such difference impacts the optimal quantization policy. 

\subsection{Quantization}

We linearly quantize the weights and activations of each layer using the action $a_k$ given by our agent, as linearly quantized model only needs fixed point arithmetic unit which is more efficient to implement on the hardware.

Specifically, for each weight value $w$ in the $k$\textsuperscript{th} layer, we first truncate it into the range of $[-c, c]$, and we then quantize it linearly into $a_k$ bits:
\begin{equation}
    \text{quantize}(w, a_k, c) = \text{round}(\text{clamp}(w, c) /s) \times s,
\end{equation}
where $\text{clamp}(\cdot, x)$ is to truncate the values into $[-x, x]$, and the scaling factor $s$ is defined as $s = c / (2^{a_k - 1} - 1)$. In this paper, we choose the value of $c$ by finding the optimal value $x$ that minimizes the KL-divergence between the original weight distribution ${W}_k$ and the quantized weight distribution $\text{quantize}({W}_k, a_k, x)$:
\begin{equation}
    c = \argmin_x \mathcal{D}_\text{KL} ({W}_k \mid\mid \text{quantize}({W}_k, a_k, x)),
\end{equation}
where $\mathcal{D}_\text{KL}(\cdot \mid\mid \cdot)$ is the KL-divergence that characterizes the distance between two distributions. As for activations, we quantize the values similarly except that we truncate them into the range of $[0, c]$, not $[-c, c]$  since the activation values (which are the outputs of the ReLU layers) are non-negative.

\subsection{Reward Function}

After quantization, we retrain the quantized model for one more epoch to recover the performance. As we have already imposed the resource constraints (latency, energy) by limiting the action space (Section \ref{sect:approach:action_space}), we define our reward function $\mathcal{R}$ to be only related to the accuracy:
\begin{equation}
    \mathcal{R} = \lambda \times (\text{acc}_\text{quant} - \text{acc}_\text{origin}),
\end{equation}
where $\text{acc}_\text{origin}$ is the top-1 classification accuracy of the full-precision model on the training set, $\text{acc}_\text{quant}$ is the accuracy of the quantized model after finetuning, and $\lambda$ is a scaling factor which is set to $0.1$ in our experiments.

\subsection{Agent}

For the RL agent, we leverage the deep deterministic policy gradient (DDPG)~\cite{Lillicrap:2016ww}, which is an off-policy actor-critic algorithm for continuous control problem. In our environment, one step means that our agent makes an action to decide the number of bits assigned to the weights or activations of a specific layer, while one episode is composed of multiple steps, where our RL agent makes actions to all layers. We apply a variant form of the Bellman's Equation, where each transition in an episode is defined as ${T}_k = ({O}_k, a_k, \mathcal{R}, {O}_{k + 1})$. During exploration, the ${Q}$-function is computed as
\begin{equation}
    \hat{{Q}}_k = \mathcal{R}_k - \mathcal{B} + \gamma \times {Q}({O}_{k + 1}, w({O}_{k + 1}) \mid \theta^{Q}),
\end{equation}
and the loss function can be approximated by
\begin{equation}
    \mathcal{L} = \frac{1}{N_\text{s}} \sum_{k=1}^{N_\text{s}} (\hat{{Q}}_k - {Q}({O}_k, a_k \mid \theta^{Q}))^2, \\
\end{equation}
where $N_\text{s}$ denotes the number of steps in this episode, and the baseline $\mathcal{B}$ is defined as an exponential moving average of all previous rewards in order to reduce the variance of the gradient estimation. The discount factor $\gamma$ is set to 1 since we assume that the action made for each layer should contribute equally to the final result. Moreover, as the number of steps is always finite (bounded by the number of layers), the sum of the rewards will not explode.

\subsection{Implementation Details}

In this section, we present the implementation details about RL exploration and finetuning quantized models.

\myparagraph{Agent.}

The DDPG agent consists of an actor network and a critic network. Both using the same network architecture: they take the state vector and the action from the last time step as inputs and feed them into two separate fully-connected layers with hidden sizes of $400$. After that, we add the two hidden vectors together and go through another two fully-connected layers with hidden sizes of $\{300, 1\}$. As for the actor network, we use an additional sigmoid function to project the output into the range of $[0, 1]$.

\myparagraph{Exploration.}

Optimization of the DDPG agent is carried out using ADAM~\cite{Kingma:2015us} with $\beta_1 = 0.9$ and $\beta_2 = 0.999$. We use a fixed learning rate of $10^{-4}$ for the actor network and $10^{-3}$ for the critic network.
During exploration, we employ the following stochastic process of the noise:
\begin{equation}
    w'({O}_k) \sim \mathcal{N}_\text{trunc} (w({O}_k \mid \theta^w_k), \sigma^2, 0, 1),
\end{equation}
where $\mathcal{N}_\text{trunc}(\mu, \sigma, a, b)$ is the truncated normal distribution, and $w$ is the model weights. The noise $\sigma$ is initialized as $0.5$, and after each episode, the noise is decayed exponentially with a decay rate of $0.99$. 

\myparagraph{Finetuning.}

During exploration, we finetune the quantized model for one epoch to help recover the performance (using SGD with a fixed learning rate of $10^{-3}$ and momentum of $0.9$). We randomly select 100 categories from ImageNet~\cite{Deng:2009td} to accelerate the model finetuning during exploration. After exploration, we quantize the model with our best policy and finetune it on the full dataset.

\section{Experiments}

We conduct extensive experiments to demonstrate the consistent effectiveness of our framework for multiple objectives: \emph{latency}, \emph{energy}, and \emph{model size}.

\myparagraph{Datasets and Models.}

Our experiments are performed on the ImageNet~\cite{Deng:2009td} dataset. As our focus is on more efficient models, we extensively study the quantization of MobileNet-V1~\cite{Howard:2017wz} and MobileNet-V2~\cite{Sandler:2018wy}.  Both MobileNets are inspired from the depthwise separable convolutions~\cite{Chollet:2017vb} and replace the regular convolutions with the \emph{pointwise} and \emph{depthwise} convolutions: MobileNet-V1 stacks multiple ``\emph{depthwise} -- \emph{pointwise}" blocks repeatedly; while MobileNet-V2 uses the ``\emph{pointwise} -- \emph{depthwise} -- \emph{pointwise}" blocks as its basic building primitives.

\begin{table}[t]
    \renewcommand*{\arraystretch}{1.15}
    \setlength{\tabcolsep}{4pt}
    \small\centering
    \begin{tabular}{lcccccc} 
        \toprule
        & Hardware & Batch & PE Array & AXI port & Block RAM \\
        \midrule
        Edge & Zynq-7020 & 1 & 8$\times$8 & 4$\times$64b & 140$\times$36Kb \\
        Cloud & VU9P & 16 & 16$\times$16 & 4$\times$256b & 2160$\times$36Kb \\
        \bottomrule
    \end{tabular}
    \vspace{-6pt}
    \caption{The configurations of edge and cloud accelerators.}
    \label{tbl:edge_cloud}
    \vspace{-12pt}
\end{table}

\subsection{Latency-Constrained Quantization}

We first evaluate our framework under \emph{latency} constraints on two representative hardware architectures: spatial and temporal architectures for multi-precision CNN. We show that it's beneficial to have specialized quantization policies for different hardware architectures. We systematically interpret the policy given by AI to guide future human designs. 

\myparagraph{Temporal Architecture.}

Bit-Serial Matrix Multiplication Overlay (BISMO) proposed by Yaman~\etal~\cite{umuroglu2018bismo} is a classic temporal design of neural network accelerator on FPGA. It introduces bit-serial multipliers which are fed with one-bit digits from 256 weights and corresponding activations in parallel at one time and accumulates their partial products by shifting over time.
 
\myparagraph{Spatial Architecture.}

BitFusion architecture proposed by Hardik~\etal\cite{sharma2018bit} is a state-of-the-art spatial ASIC design for neural network accelerator. It employs a 2D systolic array of Fusion Units which spatially sum the shifted partial products of two-bit elements from weights and activations. 

\begin{table*}[!t]
    \renewcommand*{\arraystretch}{1.15}
    \setlength{\tabcolsep}{3pt}
    \small\centering
    \begin{tabular}{lccccccccccccccc} 
        \toprule
        & & \multicolumn{6}{c}{Edge Accelerator} & \multicolumn{6}{c}{Cloud Accelerator}  \\
        \cmidrule(lr){3-8}\cmidrule(lr){9-14}

        & & \multicolumn{3}{c}{MobileNet-V1} & \multicolumn{3}{c}{MobileNet-V2} & \multicolumn{3}{c}{MobileNet-V1} & \multicolumn{3}{c}{MobileNet-V2} \\
        \cmidrule(lr){3-5}\cmidrule(lr){6-8}\cmidrule(lr){9-11}\cmidrule(lr){12-14}
        
        & Bitwidths & Acc.-1 & Acc.-5 & Latency & Acc.-1 & Acc.-5 & Latency & Acc.-1 & Acc.-5 & Latency & Acc.-1 & Acc.-5 & Latency \\
        \midrule
        PACT~\cite{Choi:2018uw} & 4 bits & 62.44 & 84.19 & 45.45 ms & 61.39 & 83.72 & 52.15 ms & 62.44 & 84.19 & 57.49 ms & 61.39 & 83.72 & 74.46 ms \\
        Ours & \emph{flexible} & \textbf{67.40} & \textbf{87.90} & 45.51 ms & \textbf{66.99} & \textbf{87.33} & 52.12 ms & \textbf{65.33} & \textbf{86.60} & 57.40 ms & \textbf{67.01} & \textbf{87.46} & 73.97 ms \\
        \midrule
        PACT~\cite{Choi:2018uw} & 5 bits & 67.00 & 87.65 & 57.75 ms & 68.84 & 88.58 & 66.94 ms & 67.00 & 87.65 & 77.52 ms & 68.84 & 88.58 & 99.43 ms \\
        Ours & \emph{flexible} & \cellcolor{red!15}\textbf{70.58} & \cellcolor{red!15}\textbf{89.77} & \cellcolor{red!15}57.70 ms & \textbf{70.90} & \textbf{89.91} & 66.92 ms & \cellcolor{blue!15}\textbf{69.97} & \cellcolor{blue!15}\textbf{89.37} & \cellcolor{blue!15}77.49 ms & \textbf{69.45} &\textbf{88.94} & 99.07 ms  \\
        \midrule
        PACT~\cite{Choi:2018uw} & 6 bits & 70.46 & 89.59 & 70.67 ms & 71.25 & 90.00 & 82.49 ms & 70.46 & 89.59 & 99.86 ms & 71.25 & 90.00 & 127.07 ms \\
        Ours & \emph{flexible} & \textbf{71.20} & \textbf{90.19} & 70.35 ms & \cellcolor{violet!15}\textbf{71.89} & \cellcolor{violet!15}\textbf{90.36} & \cellcolor{violet!15}82.34 ms & \textbf{71.20} & \textbf{90.08} & 99.66 ms & \cellcolor{olive!15}\textbf{71.85} & \cellcolor{olive!15}\textbf{90.24} & \cellcolor{olive!15}127.03 ms  \\
        \midrule
        Original & 8 bits & \cellcolor{red!15}70.82 & \cellcolor{red!15}89.85 & \cellcolor{red!15}96.20 ms & \cellcolor{violet!15}71.81 & \cellcolor{violet!15}90.25 & \cellcolor{violet!15}115.84 ms & \cellcolor{blue!15}70.82 & \cellcolor{blue!15}89.85 & \cellcolor{blue!15}151.09 ms & \cellcolor{olive!15}71.81 & \cellcolor{olive!15}90.25 & \cellcolor{olive!15}189.82 ms \\
        \bottomrule
    \end{tabular}
    \vspace{-6pt}
    \caption{Latency-constrained quantization on BISMO (edge accelerator and cloud accelerator) on ImageNet. Our framework can reduce the latency by \textbf{1.4$\times$} to \textbf{1.95$\times$} with negligible loss of accuracy compared with the fixed bitwidth (8 bits) quantization.}
    \label{tbl:bismo_latency}
    \vspace{-12pt}
\end{table*}
\begin{figure}[!t]
    \centering
    \includegraphics[width=1.05\linewidth]{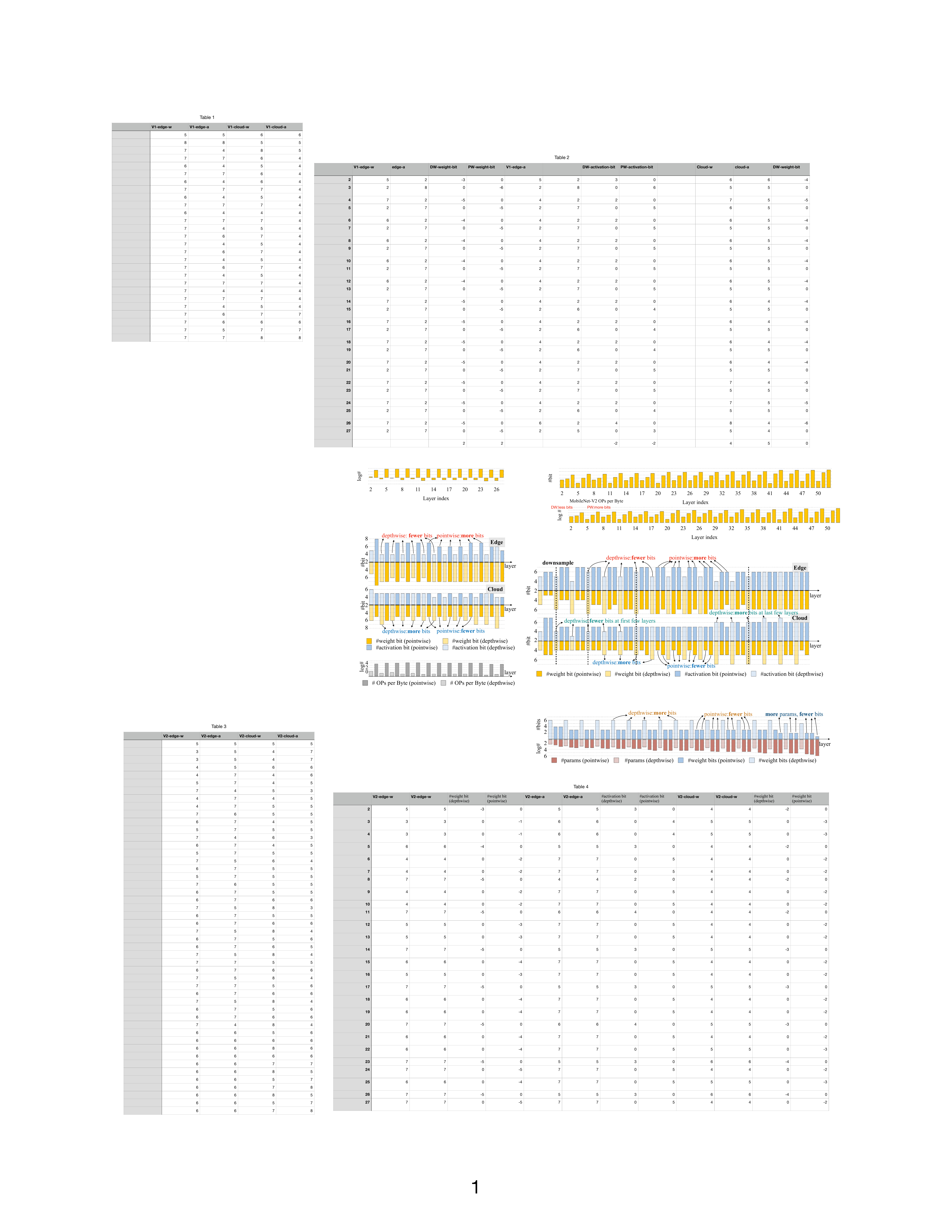}
    \caption{Quantization policy under latency constraints for MobileNet-V1. On edge accelerator, our RL agent allocates \emph{less} activation bits to the depthwise convolutions, which echos that the depthwise convolutions are memory bounded and the activations dominates the memory access. On cloud accelerator, our agent allocates \emph{more} bits to the depthwise convolutions and allocates \emph{less} bits to the pointwise convolutions, as cloud device has more memory bandwidth and high parallelism, the network appears to be computation bounded.}
    \label{fig:bismo_v1}
    \vspace{-12pt}
\end{figure}
\begin{figure*}[!t]
    \centering
    \vspace{-20pt}
    \includegraphics[width=0.93\linewidth]{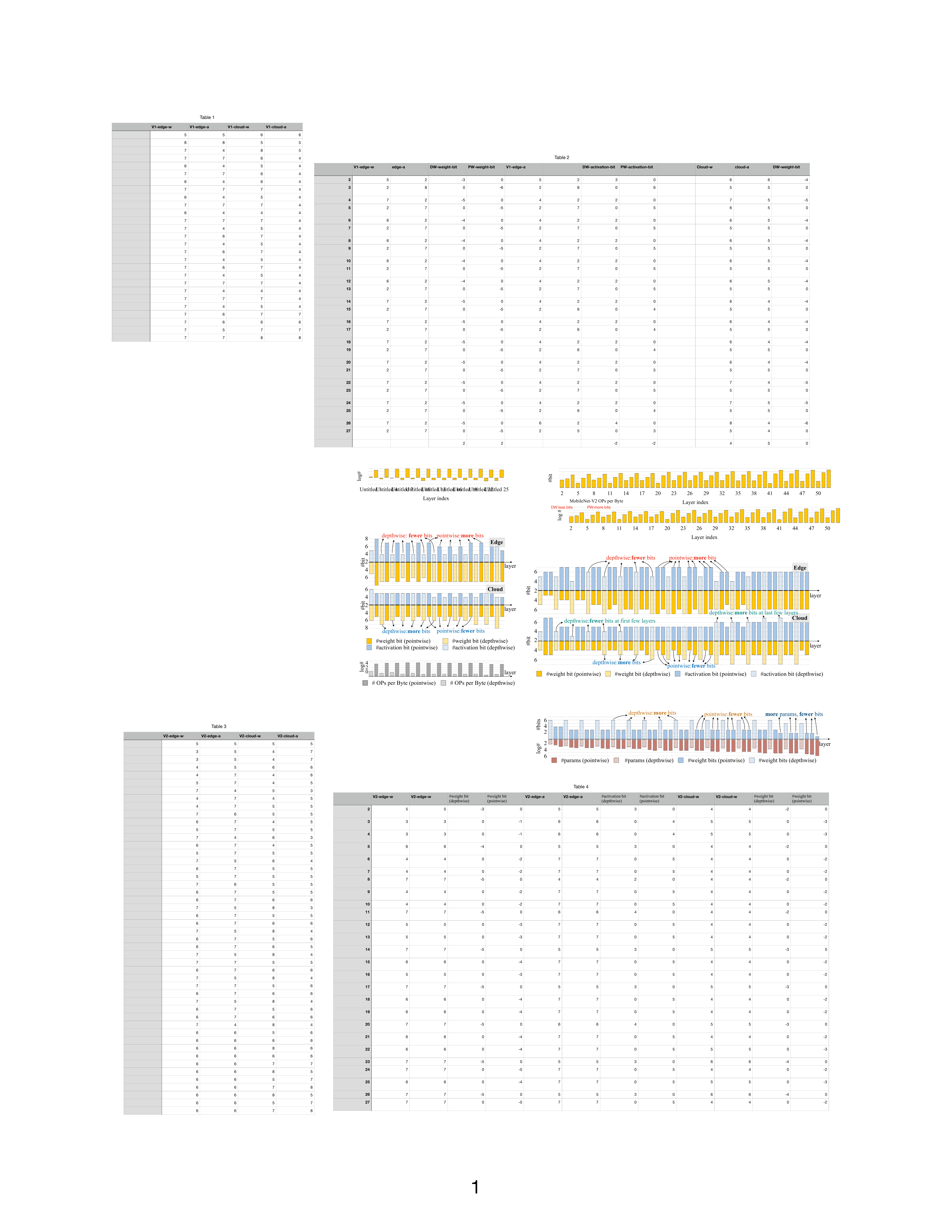}
    \caption{Quantization policy under latency constraints for MobileNet-V2 on BISMO. Similar to \fig{fig:bismo_v1}, depthwise layer is assigned with fewer bits on the edge accelerator, and pointwise layer is assigned with fewer bits on the cloud accelerator.}
    \label{fig:bismo_latency_v2}
    \vspace{-12pt}
\end{figure*}

\subsubsection{Quantization policy for BISMO Architecture}

Inferencing neural networks on edge devices and cloud severs can be quite different: batch size, memory bandwidth, peak FLOPs, \etc. We use Xilinx Zynq-7020 FPGA~\cite{zync7020} as our edge device and Xilinx VU9P~\cite{vu9p} as our cloud device. \tbl{tbl:edge_cloud} shows our experiment configurations on these two platforms along with their available resources.

As for comparison, we adopt the PACT~\cite{Choi:2018uw} as our baseline, which uses the same number of bits for all layers except for the first layer which extracts the low level features, they use 8 bits for both weights and activations as it has fewer parameters and is very sensitive to errors. We follow a similar setup for the first layer (8 bits), and explore the bitwidth allocation policy for all the other layers.
Under the same latency, \modelshort consistently achieved better accuracy than the baseline on both the cloud and the edge (\tbl{tbl:bismo_latency}). With similar accuracy, \modelshort can reduce the latency by 1.4$\times$ to 1.95$\times$ compared with the baseline.

\myparagraph{Interpreting the quantization policy.}
Our agent gave quite different quantization policy for edge and cloud accelerators (\fig{fig:bismo_v1}).
For the activations, the depthwise convolution layers are assigned less bitwidth than the pointwise layers on the edge; while on the cloud device, the bitwidth of these two types of layers are similar. For weights, the bitwidth of these types of layers are nearly the same on the edge; while on the cloud, the depthwise convolution layers got more bitwidth than the pointwise convolution layers.

We explain the difference of quantization policy between edge and cloud by the roofline model~\cite{williams2009roofline}. Many previous works use FLOPs or BitOPs as metrics to measure computation complexity. However, they are not able to directly reflect the latency, since there are many other factors influencing the hardware performance, such as memory access cost and degree of parallelism~\cite{Sandler:2018wy, liu2017learning}. Taking computation and memory access into account, the roofline model assumes that applications are either computation-bound or memory bandwidth-bound, if not fitting in on-chip caches, depending on their operation intensity. Operation intensity is measured as operations (MACs in neural networks) per byte accessed. A lower operation intensity indicates suffering more from the memory access.

\begin{table}[!t]
    \renewcommand*{\arraystretch}{1.15}
    \setlength{\tabcolsep}{4pt}
    \small\centering
    \begin{tabular}{lccccc}
        \toprule
        & Weights & Activations & Acc.-1 & Acc.-5 & Latency \\
        \midrule
        PACT~\cite{Choi:2018uw} & 4 bits & 4 bits & 62.44 & 84.19 & 7.86 ms \\
        Ours & \emph{flexible} & \emph{flexible} & \textbf{67.45} & \textbf{87.85} & 7.86 ms \\
        \midrule
        PACT~\cite{Choi:2018uw} & 6 bits & 4 bits & 67.51 & 87.84 & 11.10 ms \\
        Ours & \emph{flexible} & \emph{flexible} & \cellcolor{red!15}\textbf{70.40} & \cellcolor{red!15}\textbf{89.69} & \cellcolor{red!15}11.09 ms \\
        \midrule
        PACT~\cite{Choi:2018uw} & 6 bits & 6 bits & 70.46 & 89.59 & 19.99 ms \\
        Ours & \emph{flexible} & \emph{flexible} & \textbf{70.90} & \textbf{89.95} & 19.98 ms \\
        \midrule
        Original & 8 bits & 8 bits & \cellcolor{red!15}70.82 & \cellcolor{red!15}89.85 & \cellcolor{red!15}20.08 ms \\
        \bottomrule
    \end{tabular}
    \vspace{-6pt}
    \caption{Latency-constrained quantization on BitFusion (MobileNet-V1 on ImageNet). Our framework can reduce the latency by \textbf{2$\times$} with almost no loss of accuracy compared with the fixed bitwidth (8 bits) quantization.}
    \label{tbl:bitfusion_latency}
    \vspace{-12pt}
\end{table}

The bottom of \fig{fig:bismo_v1} shows the operation intensities (OPs per Byte) of convolution layers in the MobileNet-V1. Depthwise convolution is memory bounded, and the pointwise convolution is computation bounded. Our experiments show that when running MobileNet-V1 on the edge devices with small batch size, its latency is dominated by the depthwise convolution layers. Since the feature maps take a major proportion in the memory of depthwise convolution layers, our agent gives the activations less bits. In contrast, when running MobileNet-V1 on the cloud with large batch size, our agent increases the bitwidth of depthwise convolution to preserve the accuracy at low memory overhead since depthwise convolution only takes a small proportion of the total weights. A similar phenomenon can be observed in \fig{fig:bismo_latency_v2} on MobileNet-V2. Moreover, as the activation size in deeper layers gets smaller, they get assigned more bits.

\begin{table}[!t]
    \renewcommand*{\arraystretch}{1.15}
    \setlength{\tabcolsep}{4pt}
    \small\centering
    \begin{tabular}{lccccc} 
        \toprule
        & Weights & Activations & Acc.-1 & Acc.-5 & Energy \\
        \midrule
        PACT~\cite{Choi:2018uw} & 4 bits & 4 bits & 62.44 & 84.19 & 13.47 mJ \\
        Ours & \emph{flexible} & \emph{flexible} & \textbf{64.78} & \textbf{85.85} & 13.69 mJ \\
        \midrule
        PACT~\cite{Choi:2018uw} & 6 bits & 4 bits & 67.51 & 87.84 & 16.57 mJ \\
        Ours & \emph{flexible} & \emph{flexible} & \cellcolor{red!15}\textbf{70.37} & \cellcolor{red!15}\textbf{89.40} & \cellcolor{red!15}16.30 mJ \\
        \midrule
        PACT~\cite{Choi:2018uw} & 6 bits & 6 bits & 70.46 & 89.59 & 26.80 mJ \\
        Ours & \emph{flexible} & \emph{flexible} & \textbf{70.90} & \textbf{89.73} & 26.67 mJ \\
        \midrule
        Original & 8 bits & 8 bits & \cellcolor{red!15}70.82 & \cellcolor{red!15}89.95 & \cellcolor{red!15}31.03 mJ \\
        \bottomrule
    \end{tabular}
    \vspace{-6pt}
    \caption{Energy-constrained quantization on BitFusion (MobileNet-V1 on ImageNet). Our framework reduces the power consumption by \textbf{2$\times$} with nearly no loss of accuracy compared with the fixed bitwidth quantization.}
    \label{tbl:bitfusion_energy}
    \vspace{-12pt}
\end{table}
\vspace{-10pt}
\begin{table*}[!t]
    \renewcommand*{\arraystretch}{1.15}
    \small\centering
    \begin{tabular}{lcccccccccccc} 
        \toprule
        & & \multicolumn{3}{c}{MobileNet-V1} & \multicolumn{3}{c}{MobileNet-V2} & \multicolumn{3}{c}{ResNet-50} \\
        \cmidrule(lr){3-5}\cmidrule(lr){6-8}\cmidrule(lr){9-11}
        & Weights & Acc.-1 & Acc.-5 & Model Size & Acc.-1 & Acc.-5 & Model Size & Acc.-1 & Acc.-5 & Model Size \\
        \midrule
        Han~\etal~\cite{Han:2016uf} & 2 bits & 37.62 & 64.31 & 1.09 MB & 58.07 & 81.24 & 0.96 MB & 68.95 & 88.68 & 6.32 MB \\
        Ours & \emph{flexible} & \textbf{57.14} & \textbf{81.87} & 1.09 MB & \textbf{66.75} & \textbf{87.32} & 0.95 MB & \textbf{70.63} & \textbf{89.93} & 6.30 MB \\
        \midrule
        Han~\etal~\cite{Han:2016uf} & 3 bits & 65.93 & 86.85 & 1.60 MB & 68.00 & 87.96 & 1.38 MB & 75.10 & 92.33 & 9.36 MB \\
        Ours & \emph{flexible} & \textbf{67.66} & \textbf{88.21} & 1.58 MB & \textbf{70.90} & \textbf{89.76} & 1.38 MB & \textbf{75.30} & \textbf{92.45} & 9.22 MB \\
        \midrule
        Han~\etal~\cite{Han:2016uf} & 4 bits  & 71.14 & 89.84 & 2.10 MB & 71.24 & 89.93 & 1.79 MB & \textbf{76.15} & 92.88 & 12.40 MB \\
        Ours & \emph{flexible} & \textbf{71.74} & \textbf{90.36} & 2.07 MB & \textbf{71.47} & \textbf{90.23} & 1.79 MB & 76.14 & \textbf{92.89} & 12.14 MB \\
        \midrule
        Original & 32 bits & 70.90 & 89.90 & 16.14 MB & 71.87 & 90.32 & 13.37 MB & 76.15 & 92.86 & 97.49 MB \\
        \bottomrule
    \end{tabular}
    \vspace{-6pt}
    \caption{Model size-constrained quantization on ImageNet. Compared with Deep Compression~\cite{Han:2017td}, our framework achieves higher accuracy under similar model size (especially under high compression ratio).}
    \label{tbl:model_size}
\end{table*}

\begin{figure*}[!t]
    \centering
    \vspace{-5pt}
    \includegraphics[width=\linewidth]{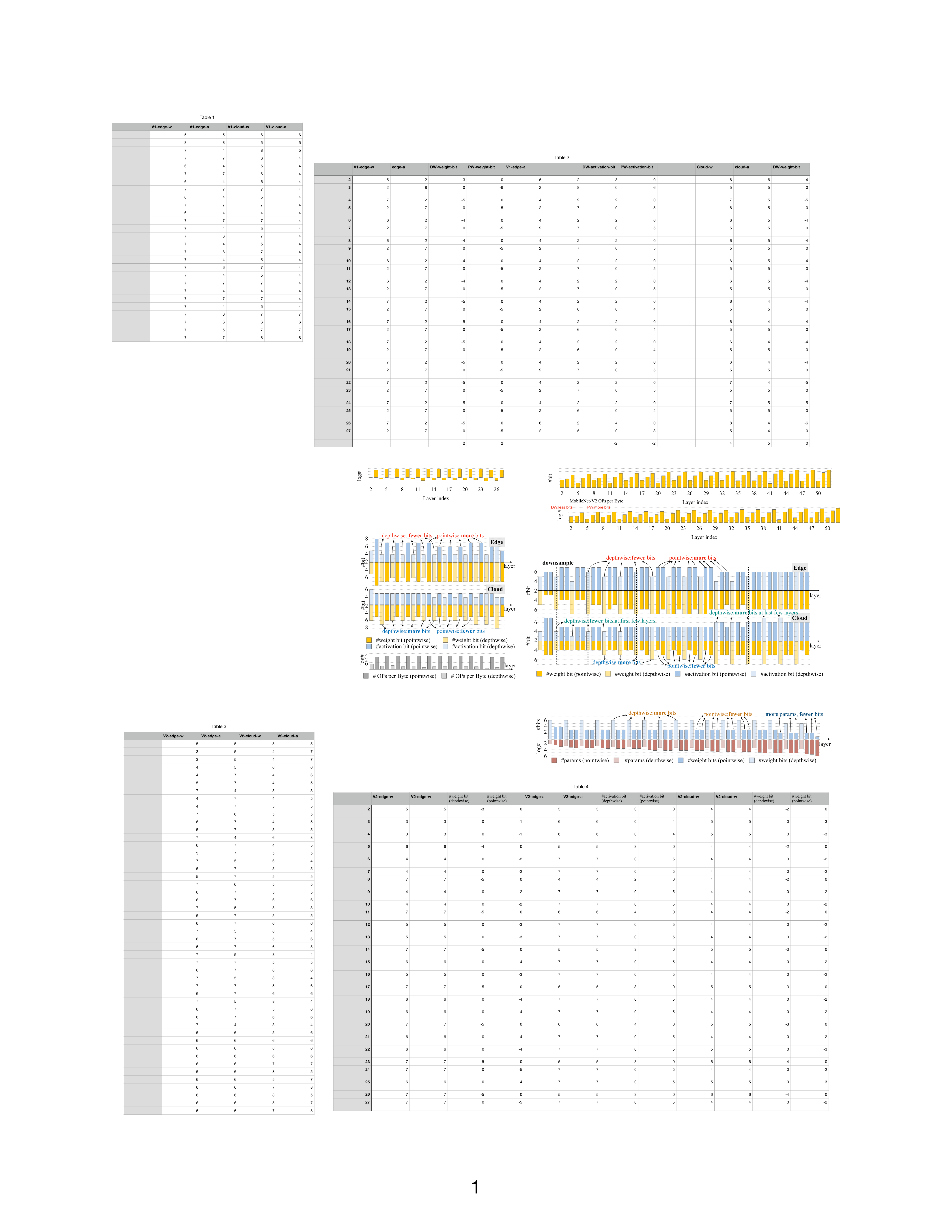}
    \vspace{-10pt}
    \caption{Quantization policy under model size constraints for MobileNet-V2. Our RL agent allocates \emph{more} bits to the depthwise convolutions, since depthwise convolutions have \emph{fewer} number of parameters.}
    \label{fig:model_size}
    \vspace{-16pt}
\end{figure*}

\subsubsection{Quantization policy for BitFusion Architecture}

In order to demonstrate the effectiveness of our framework on different hardware architectures, we further compare our framework with PACT~\cite{Choi:2018uw} under the latency constraints on the BitFusion~\cite{sharma2018bit} architecture (\tbl{tbl:bitfusion_latency}). Our framework performs much better than the hand-craft policy with the same latency. It can achieve almost no degradation of accuracy with only half of the latency used by the original MobileNet-V1 model (from \textbf{20.08} to \textbf{11.09} ms). Therefore, our framework is flexible to provide specialized quantization policy for different hardware platforms.

\subsection{Energy-Constrained Quantization}

We then evaluate our framework under the energy constraints. Similar to the latency-constrained experiments, we compare our framework with PACT~\cite{Choi:2018uw} that uses fixed number of bits without hardware feedback. From \tbl{tbl:bitfusion_energy}, we can clearly see that our framework outperforms the rule-based baseline: it achieves much better performance while consuming similar amount of energy. In particular, our framework is able to achieve almost no loss of accuracy with nearly half of the energy consumption of the original MobileNet-V1 model (from \textbf{31.03} to \textbf{16.57} mJ), which suggests that mixed precision with hardware-aware, specialized quantization policy can indeed help reduce the energy consumption.

\subsection{Model Size-Constrained Quantization}

Finally, we evaluate our framework under the model size constraints. Following Han~\etal~\cite{Han:2016uf}, we employ the $k$-means algorithm to quantize the values into $k$ different centroids instead of using the linear quantization for compression, since $k$-means quantization can be more effective reducing the model size.

We compare our framework with Deep Compression~\cite{Han:2016uf} on MobileNets and ResNet-50. From \tbl{tbl:model_size}, we can see that our framework performs much better than Deep Compression: it achieves higher accuracy with the same model size. For compact models like MobileNets, Deep Compression significantly degrades the performance especially under aggressive quantization, while our framework can preserve the accuracy much better. For instance, when Deep Compression quantizes the weights of MobileNet-V1 to 2 bits, the accuracy drops significantly from 70.90 to \textbf{37.62}; while our framework can still achieve \textbf{57.14} of accuracy with the same model size. The reason is our framework makes full use of the mixed precision by systematically searching the optimal quantization policy.

\myparagraph{Discussions.}

In \fig{fig:model_size}, we visualize the bitwidth allocation strategy for MobileNet-V2. From this figure, we can observe that our framework assigns \emph{more} bitwidths to the weights in depthwise convolution layers than pointwise convolution layers. Intuitively, this is because the number of parameters in the former is much smaller than the latter. Comparing \fig{fig:bismo_latency_v2} and \fig{fig:model_size}, the policies are drastically different under different optimization objectives (\textbf{fewer} bitwiths for depthwise convolutions under \emph{latency} optimization, \textbf{more} bitwidths for depthwise convolutions under \emph{model size} optimization). Our framework succeeds in learning to adjust its bitwidth policy under different constraints.
\section{Conclusion}

In this paper, we propose \model (\modelshort), an automated framework for quantization which does not require any domain experts and rule-based heuristics. We provide a learning based method that can search the quantization policy with hardware feedback. Compared with indirect proxy signals, our framework can offer a specialized quantization solution for different hardware platforms. Extensive experiments demonstrate that our framework performs better than conventional rule-based approaches for multiple objectives: latency, energy and model size. Our framework reveals that the optimal policies on different hardware architectures are drastically different, and we interpreted the implication of those policies. We believe the insights will inspire the future software and hardware co-design for efficient deployment of deep neural networks. 
\myparagraph{Acknowledgements.}
We thank MIT Quest for Intelligence, MIT-IBM Watson AI Lab, Xilinx, Samsung, Intel, ARM, Qualcomm, and SONY for supporting this research. We thank Google Cloud and AWS Machine Learning Research Awards for providing the computation resource.

{\small
\bibliographystyle{ieee}
\bibliography{reference}
}

\end{document}